\pdfoutput=1
\documentclass[10pt,twocolumn,letterpaper]{article}
\UseRawInputEncoding
\usepackage{wacv}
\usepackage{times}
\usepackage{epsfig}
\usepackage{booktabs}
\usepackage{cite}
\usepackage{amsmath,amssymb,amsfonts}
\usepackage{algorithmic}
\usepackage{graphicx}
\usepackage{textcomp}
\usepackage{xcolor}
\usepackage{tabularx}
\usepackage{multirow}
\definecolor{light-red}{RGB}{255, 133, 120}
\usepackage[numbers,sort]{natbib}   
\usepackage{makecell}

\usepackage[pagebackref=true,breaklinks=true,letterpaper=true,colorlinks,bookmarks=false]{hyperref}

\wacvfinalcopy 

\def\wacvPaperID{****} 

\ifwacvfinal\fi
\begin{document}

\title{A Review of Dialogue Systems: From Trained Monkeys to Stochastic Parrots}

\author{Atharv Singh Patlan$^*$
\\ 
Dept. of Computer Science and Engineering\\
IIT Kanpur\\
{\tt\small atharvsp@iitk.ac.in}
\and
Shiven Tripathi$^*$
\\ 
Dept. of Electrical Engineering\\
IIT Kanpur\\
{\tt\small shiven@iitk.ac.in}
\and
Shubham Korde$^*$
\\ 
Dept. of Electrical Engineering\\
IIT Kanpur\\
{\tt\small sskorde@iitk.ac.in}
}
\maketitle
\def\thefootnote{*}\footnotetext{All authors contributed equally to this work}\def\thefootnote{\arabic{footnote}}
\ifwacvfinal\thispagestyle{empty}\fi

\begin{abstract}
In spoken dialogue systems, we aim to deploy artificial intelligence to build automated dialogue agents that can converse with humans. Dialogue systems are increasingly being designed to move beyond just imitating conversation and also improve from such interactions over time. In this survey, we present a broad overview of methods developed to build dialogue systems over the years. Different use cases for dialogue systems ranging from task-based systems to open domain chatbots motivate and necessitate specific systems. Starting from simple rule-based systems, research has progressed towards increasingly complex architectures trained on a massive corpus of datasets, like deep learning systems. Motivated with the intuition of resembling human dialogues, progress has been made towards incorporating emotions into the natural language generator, using reinforcement learning. While we see a trend of highly marginal improvement on some metrics, we find that limited justification exists for the metrics, and evaluation practices are not uniform. To conclude, we flag these concerns and highlight possible research directions.
\end{abstract}

\section{Introduction}

There are two kinds of dialogue generation systems that are the most prevalent and in demand. The first category is task-oriented dialogue generation systems, which aim to aid the user in completing certain specific tasks like booking hotels or flights or finding details about COVID. The second category is that of Chat-bots or non-task-oriented dialogue systems, which are usually general-purpose dialogue systems that can perform a wide variety of tasks and carry out a sensible conversation with the user as well. Examples of this can include Google Assistant, Alexa, and other such AI systems.

Our paper aims to make a 2-fold contribution to the field of dialogue generation: 1) We present various dialogue generation systems developed in chronological order and 2) Highlight problems in evaluation and provide further research directions.

In our article, we present the development of dialogue generation systems in both these tasks through the ages by looking at the progress in the area through time and trying to determine how it will be in the future. For the past methods, we will be talking majorly about methods that use rules and those which draw statistical inferences from large datasets. Coming to the present, we see the prevalence of Deep Learning methods in the area, and we speculate that the future of these systems will be dominated by Deep Reinforcement learning systems with emotional intelligence, seeing the recent trends in the scientific literature tackling these problems.

Finally, we bring about the problems in the methods used to create Dialogue Generation Systems, specifically the lack of common datasets and metrics which can be used to compare these methods through the times. The absence of these makes it difficult to judge whether the methods are actually improving over the years or are they essentially the same as those that were used in the past. We also present ways in which the scientific community can approach this issue in the future so as to provide a systematic comparison of these methods, which are reliable and useful in further development in the field.

\section{Preliminaries}
\subsection{Deterministic and Statistical Methods}
\begin{enumerate}
    \item \textbf{Deterministic Rule-Based Methods}: The general principle for such methods is looking for keywords within a conversation that can be used to instruct an agent to provide predefined responses. For any such conversational agent, a 'script' can be defined from which response clauses must be generated according to expected keywords in the particular context. It is important to consider that even though the agent has a script of meaningful and intelligent responses, it has a severely limited understanding of the language itself. 
    Scripts are a set of rules which store the knowledge base of a context and act on keywords from inputs. These rules can be of the form "IF some condition THEN some action." These actions act on the input phrase by transforming, adding, or deleting symbols.
    \item \textbf{Corpus Based Statistical Methods}: In contrast with a script that contains rules with defined actions for keywords, a corpus is a collection of natural conversation. Usually, these systems are data-intensive and need to mine through huge datasets \cite{10.5555/1895550.1895693, LISON16.947} for training to generate good responses. While the rest of the paper is also focused on methods that use a corpus of datasets, here we look at methods that assume the Markov property on conversations.
    To model decision making for sequential environments with stochasticity, Markov decision processes (MDPs) \cite{LITTMAN20019240} are used. For actions taken by an agent, the environment changes state in response to it. The current state is used to determine the immediate reward obtained by the agent and the conditional probabilities for future state transitions. An agent's goal is to choose actions that maximise a long-term measure of total reward.  
\end{enumerate}
\subsection{Artificial Neural Network-based methods}
Artificial Neural Networks (ANNs) are a collection of connected computational units or nodes called neurons arranged in multiple computational layers, somewhat mimicking the human brain in the arrangement of the brain's neurons. ANN techniques are very prevalent these days in all kinds of applications, ranging from Object Detection to Voice Recognition. 

In the field of Natural Language Processing (which specifically is applied to the problem of dialogue generation), ANNs are present in some form or the other. Recurrent Neural Networks (counterparts to the feedforward neural networks used in computer vision) such as LSTM \cite{10.1162/neco.1997.9.8.1735} are a type of ANN that allow previous outputs to be used as inputs while having hidden states, thus effectively modelling a sequence of inputs. 

Hierarchical RNNs \cite{10.5555/2998828.2998898}, used by \cite{10.5555/3016387.3016435} are another type of RNNs in temporal dependencies are structured hierarchically, which allow long-term dependencies to be represented by variables with a long time scale. 

Another type of ANNs used here are Encoder-Decoder-based models, like Transformers \cite{10.5555/3295222.3295349}, which are stacks of RNNs. These use an attention mechanism that considers the context of the input, and the words related to a particular word in the input, thus giving more accurate results. In this article, we will refer to these ANN-based methods as Deep Learning methods.

\subsection{Reinforcement learning methods}
\begin{figure*}
    \centering
    \includegraphics[width=1\textwidth]{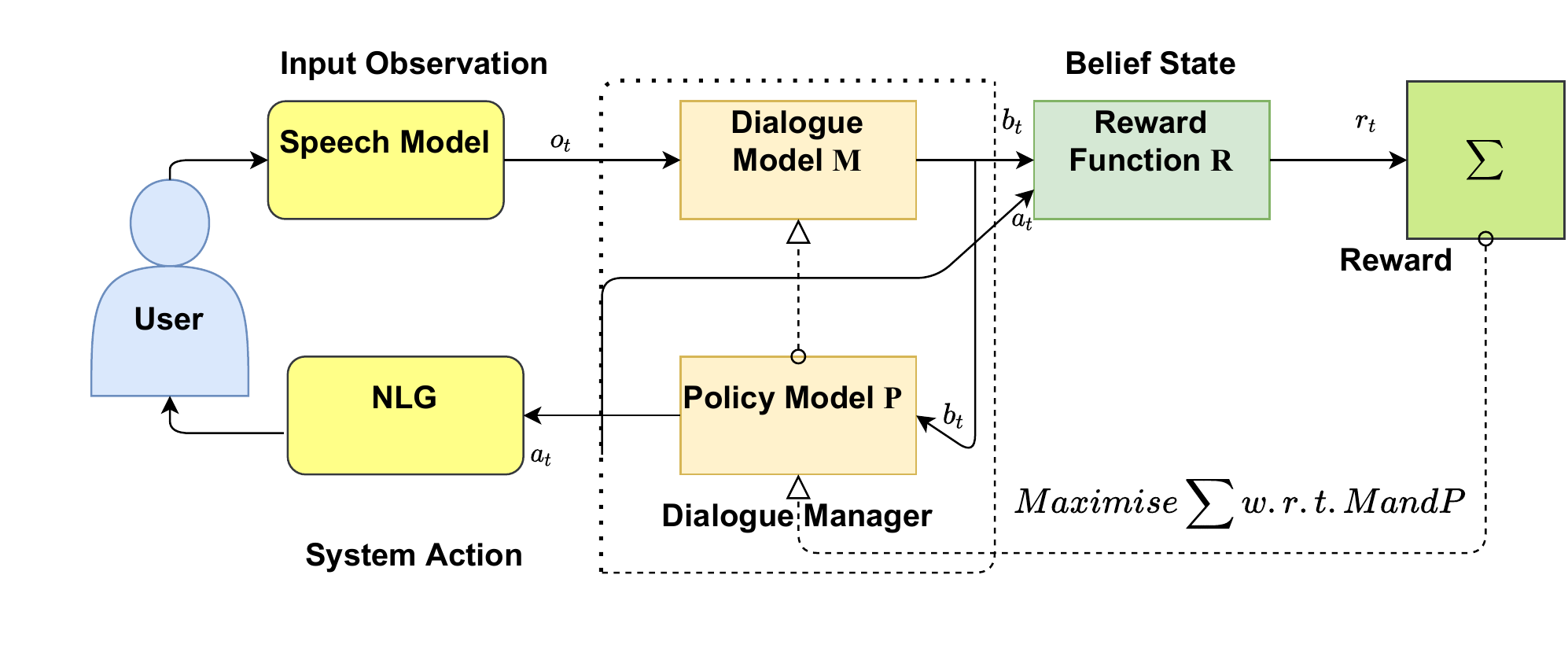}
    \caption{Representation of the pipeline followed by POMDP based spoken dialogue system.}
    \label{fig:pomdp}
\end{figure*}
Reinforcement Learning \cite{suttonbarto} is a branch of Machine Learning in which the agent learns from its actions and their consequences without any supervision. The problem is formulated as a Markov Decision Process consisting of states, actions, and rewards. The primary aim of the RL agent is to maximise the reward by finding an optimal policy.
Reinforcement Learning algorithms try to strike a balance between Exploration and Exploitation. The former refers to the ability of an agent to explore the action space. In contrast, the latter refers to its ability to exploit the gained knowledge to maximise the reward. RL algorithms are classified into three types: passive, active, and deep. In passive RL algorithms, the agent's policy is predetermined by the algorithm's designers, whereas in active RL algorithms, the agent updates its policy as it learns. Deep RL algorithms use neural networks to update the agent's policy and are most widely used. The two most common Reinforcement Learning algorithms are Q-Learning and SARSA. In both, the best estimate of future actions is used to update the current estimate. However, in the former, we take action with the greatest reward, whereas we take actual action and then calculate the reward in the latter. Reinforcement Learning is closely related to natural learning in humans and hence is highly suitable in dialogue systems because of its ability to generate a contextually sensible response.

\section{Past}
\begin{enumerate}
    \item \textbf{Rule Based Systems}:
        Designed to imitate a Rogerian psychologist (branch of psychology which draw patient out by reflecting patient's statements back at them[cite something here]), ELIZA \cite{10.1145/365153.365168} is a chatbot program created at the MIT Artificial Intelligence Laboratory by Joseph Weizenbaum. While ELIZA did not have a built-in framework to understand or contextualise events, it generated conversation by matching user input with predefined patterns and substitution methodology.

    ELIZA's success at attempting the Turing Test owes a lot to its domain choice of Rogerian psychology, which, as Weizenbaum points out, one can "assume the pose of knowing almost nothing of the real world." Most chatbots that are trying to pass the Turing test choose a domain with similar properties.

    After ELIZA, in the same clinical psychological domain, a chatbot \footnote{PARRY and ELIZA 'met' \cite{garber_2014} multiple times. At ICCC 1972, PARRY and ELIZA were connected over ARPANET, and conversed.}PARRY was developed by Colby et al. in 1971 \cite{10.1016/0004-3702(71)90002-6}, to study schizophrenia. In addition to ELIZA's pattern matching and symbol transform, PARRY included a model of its own mental state, which could change over the course of the conversation by altering 'fear' and 'anger' variables. If PARRY's anger variable is high, he will choose from a set of "hostile" outputs. If the input mentions his delusion topic, he will increase the value of his fear variable and then begin to express the sequence of statements related to his delusion. In 1972, PARRY became the first to pass the Turing test \cite{PinarSaygin2000}, as psychologists could not distinguish between transcripts of conversation generated by PARRY and those of actual paranoia patients.

    \item \textbf{Corpus Based Systems}:
    
    For dialogue systems operating in a limited domain, statistical methods have effectively been applied without the need to draft explicit rules for dialogue generation. This approach \cite{6407655, article} assumes that dialogue evolves as a Markov process, i.e., starting in some initial state, and each subsequent state is modelled by a transition probability. A dialogue model is used for representations of transit and observation probabilities. In particular, for Spoken dialogue systems, owing to the uncertainty in speech recognition, the conversation is modelled using a Partially Observed Markov Decision Process.
    A model representation is also obtained for policy, which gives us the action to take at the given state. As the dialogue progresses, a reward is assigned at each step designed to mirror the desired characteristics of the dialogue system. These desired characteristics can vary according to the system \cite{10.1016/j.csl.2006.06.008, bohus-rudnicky-2005-sorry} in use. For a system built to fill a form by asking questions to the user, a negative reward is given for more questions asked, and for a conversational system, a reward is given for longer sequences of conversation with non-trivial statements. Sometimes, offline definitions of rewards can suffer 'reward sparsity' issues due to the multitude of possible sub-steps towards the success of dialogue. Takanobu et al.\cite{DBLP:journals/corr/abs-1908-10719}, have demonstrated the applicability of adversarial training for simultaneous policy optimisation and online reward estimation.\\
    The dialogue model and policy model can then be optimised by maximising the expected accumulated sum of these rewards either online through interaction with users or offline from a corpus of dialogue s collected within a similar domain.

        \begin{figure*}
    \centering
    \includegraphics[width=1\textwidth]{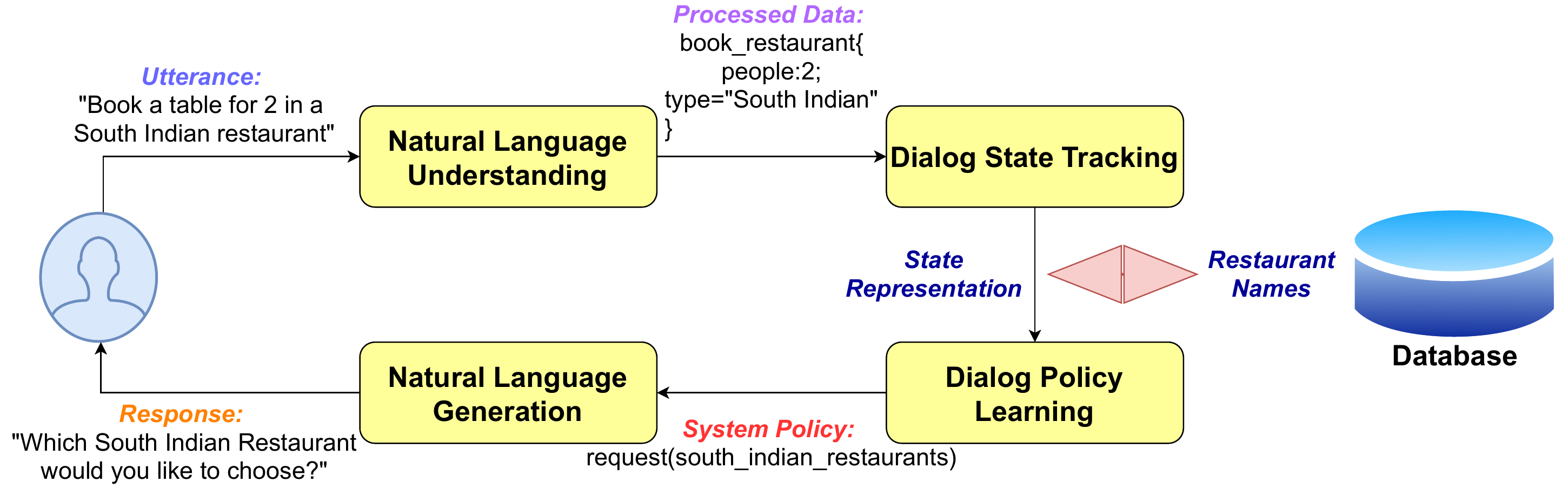}
    \caption{Representation of the pipeline followed by prevalent Task oriented dialogue systems.}
    \label{fig:endtoend}
\end{figure*}
    For offline optimisation, Value Iteration \cite{10.1016/j.csl.2006.06.008, suttonbarto} has been explored to learn optimal policies from datasets and corpus covering the necessary amount of multiple dialogue branches. After mapping the belief state (from input) to a summary feature space and taking random actions, the conditional transition function from belief feature to action feature is estimated, after which we can apply value iteration till convergence. 
    \\Using Monte Carlo optimisation \cite{10.5555/1708376.1708414}, we can estimate value function online via interaction with a user (or a simulator), with updates to value function being performed at the end of each set of dialogue s. All belief features, actions, and rewards are arranged in a tuple for each time step of dialogue. For each belief feature visited in the dialogue, updates are made to Q values using discounted returns obtained for an action and also the number of times that action was taken. An epsilon greedy strategy is utilised to explore the dialogue space for learning optimal policy. 
\end{enumerate}
\section{Present}
\subsection{Task oriented dialogue generation}
We first look at current approaches aiming to solve the task-oriented dialogue generation problem. The most prevalent approaches to solve this problem are methods that follow a particular pipeline, as shown in Figure \ref{fig:endtoend}, consisting of the following four parts: 
\begin{enumerate}
    \item \textbf{Natural Language Understanding} (NLU): NLU is responsible for identifying the intent of the user's utterance and converting it into a structured representation of predefined semantic slots. RNN methods making effective use of hidden states to model both the intent and semantic slots \cite{yao2014spoken, rnnlm, tur2016multi} are a popular approach for NLU. More complex models such as BERT \cite{devlin2019bert} can also be used to perform a similar task \cite{Chen2019BERTFJ, Castellucci2019MultilingualID}.
    \item \textbf{Dialogue State Tracking} (DST): It analyses the entire dialogue's context till a given time $t$ and tries to judge the user's end-goal at that time. The method works by maintaining a distribution over several hypotheses of the true dialogue state $H_t$, which denotes the representation of the dialogue \underline{till} $t$. This state structure formulation is called slot filling. This is used by methods such as \cite{williams-2012-belief, williams-etal-2013-dialog}, which are articles introducing a common dataset on which to evaluate the problem. Some methods such as \cite{mrksic-etal-2017-neural, Henderson2014WordBasedDS} consider the state to be composed of slot-value pairs which represent the user's end goal, and allow the problem to be modelled as a multi-class classification task, i.e., predicting $P(C_{i,t} | H_t)$, the probability of the class of the slot being $C_{i,t}$, given the dialogue state (utterances) till time $t$. The output of this module, called the \textit{state representation}, is used to retrieve relevant information, such as \texttt{Names of South Indian Restaurants} from the database, as shown in Figure \ref{fig:endtoend}.
    \item \textbf{Dialogue Policy Learning} (DPL): DPL systems are responsible for generating the next system action, which is conditioned on the dialogue state. Formulation of human conversation as a Markov Decision Process \cite{LITTMAN20019240}: at each time step, the state goes from state $s_1$ to $s_2$ by taking a particular action $a$, allows Reinforcement learning approaches to be applied to it. The current approaches use supervised learning first to train the dialogue policy off-line \cite{Yan2017BuildingTD}, and then fine-tune the model through model-free RL, with the help of DQN, such as in \cite{mnih2015dqn} or policy gradient methods, as done in \cite{Zhou2017EndtoEndOG}, using real user dialogues. Some other methods aim to use model-based RL approaches to model the environments and claim to bring down the usage of real annotated data with the help of user simulators that mimic human conversations, such as \cite{Shi2019HowTB} instead of real user dialogues. However, In addition, it is very challenging to develop a high-quality user simulator because there is no widely accepted metric to assess the quality of user simulators as noted by \cite{pietquin_hastie_2013}, and thus methods employing such simulators have not proven themselves to be of high quality. Some model-based RL methods used for DPL are \cite{Peng2018DeepDynaQ, wu2019switch}
    \item \textbf{Natural Language Generation} (NLG): NLG is responsible for mapping the system dialogue action produced by the DPL to a natural language utterance and is modelled as a conditional language generation task \cite{wen-etal-2015-semantically}. It is necessary that the generated utterances are informative, natural, specific, and similar to how the human language is, as observed by \cite{Stent2005EvaluatingEM}, and also convey the semantics of the dialogue required for task completion. Thus solving this problem as a key-value pair is not sufficient as information about the context and use-cases are needed, which can not be simply encoded as keys. Thus deep models are used to learn these representations. Along with RNN-based models such as \cite{Wen2015StochasticLG, wen-etal-2015-semantically}, models using pre-trained GPT on large-scale NLG corpus and then fine-tuning on task-specific NLG tasks are used to prepare models for this task, which are then fine-tuned to the specific task. \cite{peng-etal-2020-shot} is one such model.
\end{enumerate}

At present, methods bypassing the need of this pipeline and which implement an end-to-end system for dialogue Generation also do exist. A few methods like \cite{wen-etal-2017-network} modelled all the components of the pipeline as neural networks and combined them to perform end-to-end training. Other methods such as \cite{bordes2016learning, madotto-etal-2018-mem2seq} consider the task-oriented dialogue as a comprehension problem by taking the user's utterance as questions and the system's responses as the answers while taking the dialogue history as the overall context. It is important to note that these methods require supervised training requiring large amounts of labelled training data, which is expensive to annotate manually. They also suffer from domain variance, as they observe only some examples in the data and can not fully explore the state-action space.
\subsection{Chat-bots}

Non-task-oriented dialogue systems are mostly used for general conversations with humans, and hence it is difficult to come up with a specific pipeline for this task. Thus, this problem is implemented in two flavours:
\begin{enumerate}
    \item \textbf{Generative Methods}: Neural Generative Models used for this task follow the structure of sequence to sequence models, which are implemented using Encoder-Decoder structures, wherein the encoder converts the input sequence $X = (x_1, x_2, \dots x_N)$ consisting of $N$ words into a context vector $c$, dependent on the hidden states of the previous timestep in the encoder. $c$ is calculated by reading $X$ word to word and calculating the hidden state at time step $t$ as $h_t = f(x_t, h_{t-1})$, where $f$ is the RNN, and at the end, $c$ is assigned as $h_N$, the hidden state of the last word. On the other hand, the decoder estimates the probability of generation of the output $Y = (y_1, y_2, \dots, y_{N^{'}})$ with $c$ as input, which is conditioned on the hidden states of the decoder, in a similar fashion as the encoder. \cite{luong-etal-2015-addressing, shang-etal-2015-neural} are some examples implementing such a model. \cite{meenapaper} is a recent method using a seq2seq model along with the evolved transformer \cite{So2019TheET}, making use of a huge number of parameters (2.6 B), and is simply trained to minimise perplexity scores. However, certain sub-problems need to be tackled while implementing this method, such as taking into account the context of the dialogue and making sure that the responses generated are meaningful and diverse (and not generic like \textit{I don't know}, or \textit{OK}, which most NLG models do).
    
    \cite{10.5555/3298023.3298047} aims to tackle both these problems. It makes use of the hierarchical dialogue modelling employed by \cite{10.5555/3016387.3016435} which captures the meaning of individual utterances of the user and integrates them as conversations, which account for the context of the input, and adds stochastic latent variables to the hierarchical framework, which introduces diversity in the output generation by introducing a very small degree of randomness in the output (which they claim is kind of similar to humans as different humans might reach differently to different questions). However, the latent variable is essentially shallow and designed only to make high-level decisions about the response, such as topic or sentiment, and not the minute details.
   \item \textbf{Retrieval Based Methods}: These methods take into consideration the inputs, both at a given timestep and the ones before it and try to determine a response that best fits the context of the inputs. This makes the identification of relevant information extremely essential so that the conversation remains consistent, and very slight deviations do not cause the conversations to shift altogether. This methodology is called Multi-turn Response Matching. It is important for models in this setting to identify important information in previous utterances and properly model the utterances relationships to ensure conversation consistency. The most prevalent methods in this category usually involve the concatenation of the input context with the response or some other metric before being passed through an RNN, as done in \cite{10.1145/2911451.2911542, zhou-etal-2016-multi}.
\end{enumerate}

\section{Future}
\begin{figure}
    \centering
    \includegraphics[width=0.5\textwidth]{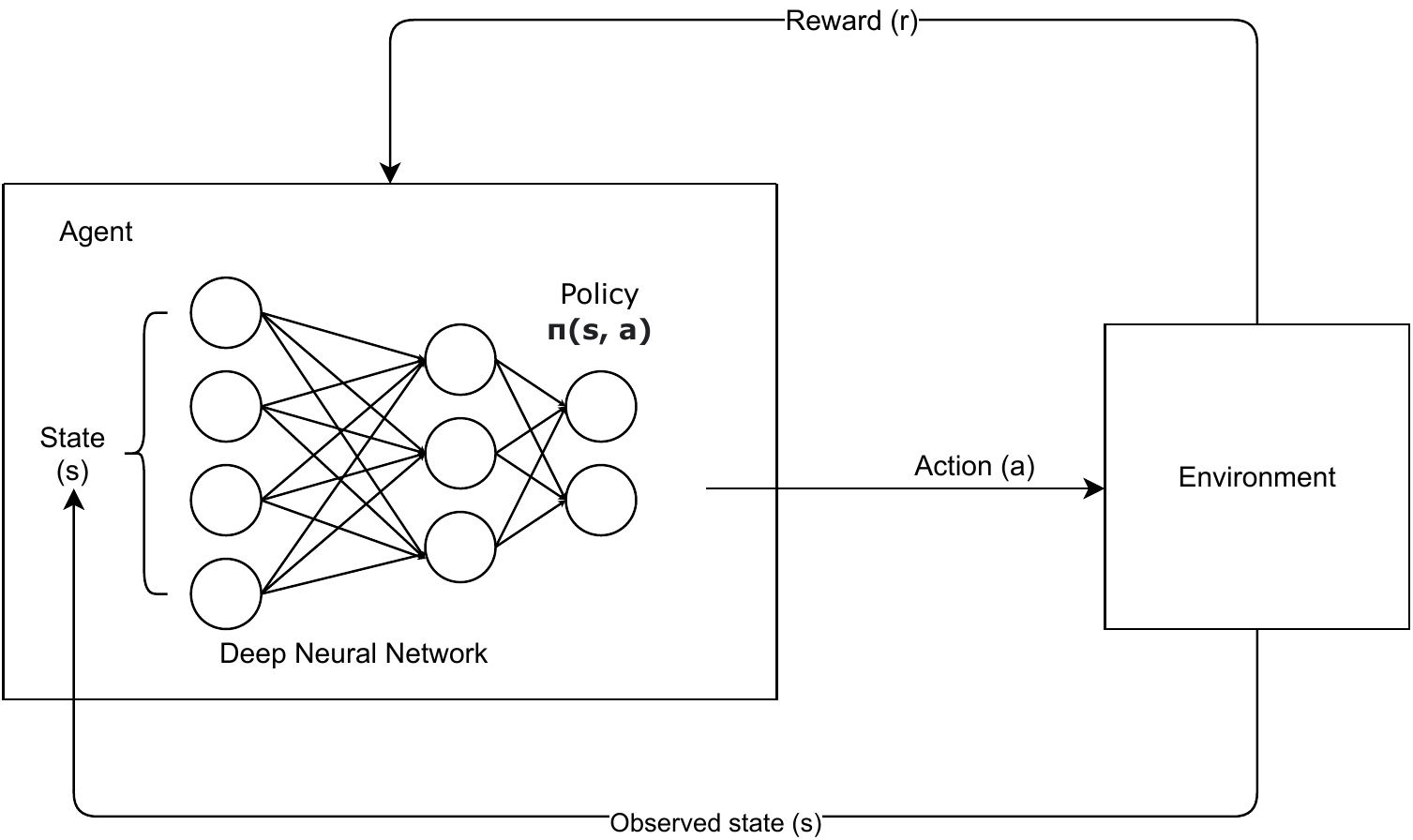}
    \caption{Architecture of Deep Reinforcement Learning}
    \label{fig:RL}
\end{figure}
\subsection{Deep Reinforcement Learning based Methods}
The conventional Seq2Seq based conversational agents generate responses one at a time, resulting in non-interesting and repetitive loops. Deep Reinforcement Learning techniques model future rewards and perform long-term goal optimization to generate coherent and interesting dialogues. 
\cite{jiweipaper}
The model is trained by simulating a conversation between two agents. The long term rewards are designed by the system developers which are based upon the factor such as coherence, interactiveness, and informativeness of the conversation. The agent tries to maximise the reward by exploring the space of possible actions. The action space includes the dialogue s that the agent can generate, and the state space includes the dialogue history. The agent's policy takes the form of an LSTM encoder-decoder with a stochastic representation. \par
The rewards are calculated using models for representing the ease of answering, information flow, and semantic coherence. The reward for ease of answering is measured by the negative log-likelihood of responding with a dull or non-interesting response from a list of such responses. The reward for information flow is incorporated in the reward function by penalising similar consecutive responses. The penalty is calculated by the negative log of the cosine similarity between the consecutive encoder representations. Grammatically incorrect or non-coherent responses are avoided by the mutual information between the generated response and the previous responses. Its reward is calculated using the forward (generating a response given previous responses) and the backward (previous responses given the generated response) probabilities of dialogue generation and scaling them by their target lengths. The final reward function is a weighted sum of these three rewards. \par

In other approaches, RL is used after the Supervised Learning systems start operating at scale, interacting with a large number of users \cite{policygradient}. An optimisation is done using policy gradient-based RL in which after every dialogue, the reward is calculated, and the gradients of the probabilities of the actions are updated wrt model weights. In this method, better responses produce higher rewards and a positive gradient resulting in an increased likelihood of the action. In contrast, poor responses produce low rewards and a negative gradient resulting in decreased likelihood of the action. Attempts have also been made to apply second-order gradient estimates like natural gradient for faster convergence \cite{naturalpolicygradient}. \par

\subsection{Methods incorporating Emotions}
The responses generated by conversational agents are grammatically correct but are often generic and non-specific. Scientists believe that the addition of an emotional aspect will improve the ability of robots and AI systems to interact with humans. The human-like aspect of conversational agents will have many applications in psychological counselling and open domain robotics. 

The future of dialogue generation lies in developing open-domain conversational agents that could identify emotions and generate emotionally intelligent responses. Although various attempts have been made 
to incorporate emotions into RL-based models for the purpose of artificial general intelligence. These methods are also valid for dialogue generation and can be broadly classified as follows:
\begin{enumerate}
    \item \textbf{Reward Shaping}  Reward Shaping is the most widely used method, involves adding a new term depending on the internal emotional state to the external reward function \cite{ngharadapaper}. The additional term is either function of the increase in the well-being \cite{salichsmalfaz}\cite{gadanho} or a decrease in the drive of the agent. If the agent wants to express an emotion of sadness, responses such as "I feel so helpless" or "I feel sorry for you" will reduce the drive of the agent to express sadness and hence, have higher rewards.
    \item \textbf{State Modification} Emotions have been embedded as part of the state-space by many researchers. The action taken by the agent is made a direct function of the emotional variable. For Ex: The robot Maggie \cite{robotmaggie} takes on fear-specific actions when the emotional state represents fear. Attempts have also been made to implement the Bayesian Affect Control Theory \cite{HoeySchroeder} using a Partially Observable Markov Decision Process (POMDP). In the Bayesian Affect Theory, the self is considered to be multi-modal and learnable, which is represented by a probability distribution. It includes a three-dimensional emotional space represented by arousal, valence, and control.  

    \item \textbf{Meta Learning}
    Researchers \cite{Doya2000MetalearningNA}have found that certain neuromodulators are closely related to parameters in reinforcement learning. The action of neuromodulators which is mostly governed by their concentration and reuptake is over a time scale ranging from a few seconds up to 20 mins \cite{dopaminepaper}. For instance, the discount factor ($\gamma$) and serotonin are closely connected. Low levels of serotonin in humans are associated with impulsive actions similar to an RL agent with low $\gamma$ concerned only with immediate rewards. Another such example is between the temporal difference error ($\delta$) and dopamine. In an experiment on monkeys, it was found after training that dopaminergic neurons responded to a cue of light instead of the actual reward, similar to TD error $\delta$ in TD-Learning. Emotions such as anger, joy, and fear have been associated with meta parameters like $\beta$, $\alpha$, and $1-\gamma$, respectively. 
    
\end{enumerate}

\begin{table*}[]
    \centering
    \begin{tabular}{|c|c|ccccccc|}
    \hline
         Model & Dataset & BLEU & \makecell{Dialogue\\Length} & \makecell{Response\\Diversity} & \makecell{Entity\\Matching} & Perplexity & \makecell{Human\\Eval.} & Others\\
    \hline
        \makecell{Network based\\End-to-End Task\\Oriented system\cite{wen-etal-2017-network}} & $\mathcal{S}$ & \checkmark & & & \checkmark & & \checkmark & \makecell{Objective Task\\Success rate}\\
    \hline
        Mem2Seq \cite{madotto-etal-2018-mem2seq} & \cite{bordes2016learning, Henderson2014WordBasedDS, eric-etal-2017-key} & \checkmark & & & \checkmark & & &\\
    \hline
        NMT for RWP\cite{luong-etal-2015-addressing} & \cite{luong-etal-2015-effective} & \checkmark &&&&&& \\
    \hline
        \makecell{Neural Responding\\Machine for\\Short-text Convo \cite{shang-etal-2015-neural}} & $\mathcal{S}$ &&&&&& \checkmark &\\
    \hline
        \makecell{Hierarchical LVM\\Encoder-Decoder \cite{10.5555/3298023.3298047}} & \cite{ritter-etal-2010-unsupervised} &&&&&& \checkmark &\\
    \hline
        \makecell{Generative\\Hierarchical NN \cite{10.5555/3016387.3016435}} & $\mathcal{S}$ &&&&& \checkmark &&\\
    \hline
        Meena-bot\cite{meenapaper} & $\mathcal{S}$ &&&&&\checkmark & \checkmark & SSA\\
    \hline
        DL2R \cite{10.1145/2911451.2911542} & $\mathcal{S}$ &&&&&&& \makecell{p@1, nDCG,\\MAP, MRR\\(see \cite{10.1145/2911451.2911542})}\\
    \hline
        Seq2Seq \cite{jiweipaper} & \cite{lison-tiedemann-2016-opensubtitles2016} & \checkmark & \checkmark & \checkmark &&&\checkmark&\\
    \hline
        \makecell{Mutual Information\\Seq2Seq \cite{jiweipaper}} & \cite{lison-tiedemann-2016-opensubtitles2016} & \checkmark & \checkmark & \checkmark &&&\checkmark&\\
    \hline
        \makecell{RLDGS \cite{jiweipaper}} & \cite{lison-tiedemann-2016-opensubtitles2016} & \checkmark & \checkmark & \checkmark &&&\checkmark&\\
    \hline
    \end{tabular}
    \caption{Comparison of the various metrics and datasets used by different dialogue Generation models. The datasets used are represented by $\mathcal{S}$ representing self collected datasets, and citations in the column represent the original source of the collected data. Checkmarks (\checkmark) under the metric names represent that the corresponding metrics has been reported by the model.}
    \label{tab:metrics}
\end{table*}
\section{Evaluation}
Evaluation of dialogue systems is still an active research field because of the multiplicity of possible acceptable responses an agent could give. For each purpose (task-oriented or conversational dialogue systems), we define metrics that are expected to show correlation with the fulfilment of the purpose. For example, a model designed to answer questions could be evaluated for correctness, which might not be a good indicator of the success of conversational agents. In the following sections, we will focus on the evaluation of Natural Language Generation (the last subsystem in dialogue generation).

\begin{enumerate}
    \item \textbf{Human Evaluation}:
    For an NLG system designed for any purpose, the end goal is to generate dialogue s that are meaningful to people. Therefore human evaluation is viewed as the most important benchmark for any system and is also used to identify reasonable automatic evaluation metrics. Evaluation is conducted by either scoring each individual result (point-wise) or comparing two competing results (pair-wise). Human evaluation can also be conducted intrinsically \cite{DBLP:journals/corr/abs-1909-01388} by scoring the quality of the generated text or extrinsically \cite{li2020results}, by judging performance and user experience in using the dialogue system for a downstream task.
    \item \textbf{Automatic Evaluation}:
    Human evaluation is problematic because of being costly, time-taking and sometimes not reproducible, prompting research in automatic evaluation which are cheap, fast, and reproducible because of their objectivity.
    \begin{enumerate}
        \item \textbf{Intrinsic Metrics}: These metrics depend on the inherent properties of the dialogues themselves, like diversity and length, and are good indicators for performance for conversational agents. The length of dialogue represents the ability of a model to have interesting conversations. A dialogue is considered to have ended when the agent generates a dull response like "I don't know" or the consecutive responses are the same. The diversity of responses is gauged by counting the number of distinct unigrams and bigrams that are generated responses. The number of unigrams and bigrams is scaled by the total number of generated tokens in order to prevent favouring long sentences. New metrics such as SSA (Sensibility and Specificity Average) \cite{meenapaper} are used to evaluate the human-like aspects of multi-turn conversations for open-domain conversational agents.
        \item \textbf{Word Overlap Metrics}: 
        For generative models, metrics such as perplexity and BLEU scores are widely used \cite{takanobu2020goaloriented}. A problem with metrics based on MAP outputs like BLEU scores is that they favour models that generate the exact pronouns and punctuation as reference data. Since this biases the response generation to reference data without justifying any correlation to the appropriateness, these metrics are not always suitable for evaluation in domains with multiple plausible answers.
        \item \textbf{Trained Metrics}:
        These metrics have been proposed recently, and initial results show potential for improvement over other automatic metrics in resemblance with human judgment. Lowe et al. \cite{lowe-etal-2017-towards} propose an Automatic dialogue Evaluation Model (ADEM), based on an RNN, trained to predict ratings given by human judges. They found Pearson's correlation with Human Judge Ratings for ADEM lie at 0.41 on the utterance level and at 0.954 on the system level, while for BLEU, the correlation values are 0.062 and 0.2688, respectively, demonstrating the better performance of trained metrics at faithfully indicating success on human evaluation.
    \end{enumerate}
\end{enumerate}
As can be seen in table \ref{tab:metrics}, every method that has been reported by a different paper uses a different dataset, either from another source or via self-collection, which makes it difficult to compare the results as the difficulties of the different datasets can not be directly compared. Moreover, different papers report results using different metrics, which further brings down the credibility of the results and consistency in comparison. The most common metric observed is Human Evaluation, and that itself is not of only one type, but can vary from task to task, such as ranking various models, or giving ratings to responses, etc., which also make it difficult to compare results reported in the literature. The second most common metric is the BLEU score, and we have already presented the issues with it above. Thus, we believe that methods to fairly compare the performance of the dialogue Generation System do not exist in the present and is also an issue the future work needs to tackle.

Some sample conversations generated by the methods in table \ref{tab:metrics} have been provided in the Appendix.

\section{Conclusion and Discussion}
In this paper, we have contributed a review of work done in developing dialogue systems, starting from the early rule-based systems to current research focussing on large-scale neural models for language generation. We covered different tasks such dialogue systems could be put to work at ranging from open domain settings like conversation to task-based modules. We also reviewed the ongoing research incorporating emotional intelligence to conversational agents. Lastly, we critically examined methods for evaluation of such systems and pointed out limitations in the current understanding of the performance of a dialogue system. The following subsections conclude the paper by identifying them along with suggesting possible research directions and recommendations for the community.

An exponential increase in computational power over the years has contributed to the exploration of more complex architectures like neural models in NLP. Most of the progress of these systems on leaderboards for specific tasks has been achieved by training an increasingly large number of parameters on huge datasets. While a yearly increase by few percentage points on a non-uniform metric with the unproven utility on a non-standardised leaderboard may seem like progress in the field, the authors want the readers to introspect about the costs associated with such systems and whether large language models are really the only research direction to pursue.

\begin{enumerate}

    \item \textit{Automatic Evaluation}:
    Metrics like BLEU, ROUGE, etc., which rely on word overlap with some reference data, have been shown to have a very low correlation with actual human judgment. To ensure that research applications are actually meeting our goals, it is desired that performance on metrics should be a strong indicator for desired characteristics of the system. Unfortunately, the inadequacy of automatic evaluation metrics has led to most authors using their own Human Evaluation to demonstrate model performance. Research could be done towards building more reasonable metrics for automatic evaluation since human metrics have been found to also suffer from the problem of reliability because of lack of reproducibility. A possible line of research that has been explored in this is Automatic Trained Metrics, which have shown promising preliminary results.
    
    \item \textit{Common Leaderboard}:
    While there have been multiple attempts using various methods for dialogue generation, we still lack papers conducting systematic comparative studies of algorithms, picking few benchmark problems, and reporting results. The creation of the benchmark dataset of ImageNet\cite{5206848} spurred research into visual recognition. Similar benchmark datasets could be developed, and leaderboards be set up by the NLP community to objectively judge performances of different approaches.
    
    \item \textit{Reward Signals}:
    Deep Reinforcement Learning's successful application is non-trivial due to difficulty in modeling environment interactions, no definite answer to the granularity of decision making needed, and hard to define reward signals. In the case of open-ended dialogue systems, little work has been dedicated to the problem of defining and predicting reward signals. Reward prediction is an intuitive way of measuring the success of our system, in which we can encode signals for task success and quality of conversation. While previous works have achieved progress for task-based systems, in which rewards can be approximated to be proportional to success at subtasks (at specific levels of granularity), future work could focus on solving the problem for the open-ended system by using methods like online estimation of rewards.\cite{DBLP:journals/corr/abs-1908-10719}
    
    \item \textit{Empirical rigour in evaluation}: At NIPS 2017, Rahimi and Recht \cite{irsog1_2017} claimed that lack of rigour in presenting results or justifying methods has led to "machine learning becoming alchemy." In particular, for deep learning systems, due to the inherent black-box nature, which limits interpretability, findings show that a lot of the research progress made could be accounted simply to authors measuring their newer model (nearly always using more computation than the previous) with weaker baselines. From our evaluation study, we found that authors have liberally used unfair baselines (\cite{jiweipaper} compares the performance of 2019 DRL systems with 2015 baselines in DL, calling it a win) to prove superior performance of their systems at a particular task. Furthermore, non-uniformity in evaluation across systems disincentivises authors to take the effort to fine-tune previous baselines, before comparing to their result. To prevent such false appearance of making a breakthrough in the field every other day, we propose productive changes for empirical rigour, with recommended strategies for evaluation to be followed uniformly like tuning methodology, layer slice analysis, ablation studies, among others.
    
\end{enumerate}

Work on dialogue Systems has been growing enormously and rapidly. Exciting new algorithms, architectures, and their applications are being made with new breakthroughs made much more often. As a consequence, this literature review is incomplete in the sense of both widths of methods and depth of evaluations. However, we hope that our attempt to summarise findings and achievements in this amazing field would help provide potential research directions and hopefully draw attention to some challenges which must be addressed before any system can claim to declare victory over the task of Dialogue Generation!



{\small
\bibliography{egbib}
}

\section{Appendix}
\begin{table*}[ht]
    \centering
    \begin{tabular}{ |p{5cm}|p{5cm}|p{5cm}|  
    }
 \hline
 \textbf{Input Message} & \textbf{Mutual Information Model} & \textbf{RL based Model}\\
 \hline
 How old are you?& I'm 16. & I'm 16, why are you asking?\\
 \hline
 What's your full name? & I have no idea. & What's yours?\\
 \hline
 I don't want to go home tonight. & Really? & Why?\\
 \hline
Do you have any feelings for me? & I don't know what you are talking about. & I don't want to hurt your feelings. \\
\hline
How much time do you have here? & Not long enough. & Sorry, sir. Ten seconds.\\
\hline
Shall we get started? & Of course! & Yes. We've got a lot of work to do here.\\
\hline
Do you play football? & No, i don't & Yes. I love football! \\
\hline
We'd have to talk to him. & I mean, he's a good guy & About what ?\\
\hline
How come you never say it? & Because I don’t want to hurt you. & I don’t think it’s a good idea to say it.\\
 \hline
\end{tabular}
    \caption{Comparison between single responses of the baseline Mutual Information Model and RL based Model. \cite{jiweipaper}}
    \label{tab:my_label}
\end{table*}

\begin{table*}[ht]
    \centering
    \begin{tabular}{ |p{7.5cm}|p{7.5cm}|  
    }
 \hline
 \textbf{Baseline mutual information model} & \textbf{ Reinforcement learning model} \\
 \hline
 A: Where are you going? (1) & A: Where are you going? (1)\\
B: I'm going to the restroom.(2) & B: I'm going to the police station. (2)\\
A: See you later. (3) & A: I'll come with you. (3)\\
B: See you later. (4) & B: No, no, no, no, you're not going anywhere. (4)\\
A: See you later. (5)& A: Why? (5)\\
B: See you later. (6)& B: I need you to stay here. (6)\\
... & A: I don't know what you are talking about. (7)\\ 
\hline
A: how old are you? (1) &A: How old are you? (1)\\
B: I'm 16. (2) &B: I'm 16. Why are you asking? (2)\\
A: 16? (3) & A I thought you were 12. (3)\\
B: I don't know what you are talking about. (4) &B: What made you think so? (4)\\
A: You don't know what you are saying. (5) &A: I don't know what you are talking about. (5)\\
B: I don't know what you are talking about. (6) &B: You don't know what you are saying. (6)\\
A: You don't know what you are saying. (7) &...\\
... & ...\\
 \hline
\end{tabular}

    \caption{dialogue Simulation between two agents based on Mutual Information based model and RL based model. \cite{jiweipaper} }
    \label{tab:my_label}
\end{table*}

    \begin{verbatim}
Human: Men are all alike.
ELIZA: IN WHAT WAY
Human: They're always bugging us about 
       something or other.
ELIZA: CAN YOU THINK OF A SPECIFIC 
       EXAMPLE
Human: Well my boyfriend made me come 
       here.
ELIZA: YOUR BOYFRIEND MADE YOU COME 
       HERE
Human: He says I'm depressed much of 
       the time.
ELIZA: I AM SORRY TO HEAR YOU ARE 
       DEPRESSED
    \end{verbatim}
\begin{center}
        A sample conversation with rule based conversational agent ELIZA. \cite{10.1145/365153.365168}
\end{center}

    \begin{verbatim}
ELIZA:  Good Evening. Tell me your
        problems. Please terminate 
        input with a period or a 
        question mark.
PARRY:  People get on my nerves 
        sometimes.
ELIZA:  I am not sure I understand 
        you fully.
PARRY:  You should pay more attention.
ELIZA:  Suppose you should pay more 
        attention.
PARRY: You're entitled to your own 
        opinion.
ELIZA:  What makes you think I am 
        entitled to my own opinion?
PARRY:  It bothers me just to be 
        around people in general.
ELIZA:  Please go on.
PARRY:  I went to the track at 
        Bay Meadows a while back.
ELIZA:  What does that suggest to 
        you?
PARRY:  I went to the races.
    \end{verbatim}
\begin{center}
        Sample conversation between PARRY and ELIZA\cite{10.1145/365153.365168}
\end{center}

    \begin{verbatim}
Human :   Restaurant in any area that 
          serves Vietnamese food
Machine : thanh binh serves vietnamese 
          food type , is there anything 
          else i can help you with ?
Human :   What is the phone number and 
          post code
Machine : The number is 01223 362456 and 
          the area code is c.b 3, 0 a.f.
Human :   Good bye
Machine : Thank you for using the 
          cambridge restaurant system .
    \end{verbatim}
\begin{center}
    A task oriented end-to-end dialogue system conversation.
    \cite{wen-etal-2017-network}
\end{center}

\begin{verbatim}
Human:  Hi!
Meena:  Hi! How are you this fine 
        evening?
Human: I'm doing well. it's morning my 
        time!
Meena:  Good morning! What are your plans
        for today?
Human:  Get my work done then watch a 
        movie at home
        in the night
Meena:  Sounds fun! What movie?
Human:  I haven't decided yet, perhaps 
        an Oscar nominated one. Any 
        suggestion?
Meena: I'm not too familiar with recent
        Oscar winners,
        but I do know I wanted to see   
        The Grand Budapest Hotel.
Human:  ah, tell me about that movie!
Meena: It's from the same director and     
        writer as Moonrise Kingdom. Wes 
        Anderson is pretty well known 
        for making funny, colorful films. 
        I've heard it's pretty good.
Human:  ha! I've searched on the internet 
        and both movies are very colorful!
Meena: I'm looking forward to it. I love 
        colorful movies.
\end{verbatim}
\begin{center}
    Sample conversation with MEENA bot. \cite{meenapaper}
\end{center}

\begin{verbatim}
DRIVER :    Where can I get tea?
ASSISTANT:  The nearest tea shop is Palo 
            Alto Cafe located 4 miles 
            away, would you like 
            directions there?
ASSISTANT:  Palo Alto Cafe is 4 miles 
            away and serves coffee and 
            tea.Do you want the address?
\end{verbatim}
\begin{center}
    Sample generated Response of in-car assistant (Mem2Seq).
    \cite{madotto-etal-2018-mem2seq}
\end{center}

\end{document}